\documentclass[letterpaper, 10 pt, conference]{ieeeconf}  % Comment this line out if you need a4paper

\IEEEoverridecommandlockouts                              % This command is only needed if 
\title{\LARGE \bf
Explain What You See: \\ 
Open-Ended Segmentation and Recognition of Occluded 3D Objects
}

\author{H. Ayoobi $^{1, 2}$,
        H. Kasaei$^{2}$, % <-this % stops a space
        M. Cao $^{2}$,
        R. Verbrugge$^{2}$,
        B. Verheij$^{2}$
\thanks{$^{1}$ Computing Department, Faculty of Engineering, Imperial College London,
{\tt\small h.ayoobi@imperial.ac.uk}}
\thanks{$^{2}$ Bernoulli Institute,
        University of Groningen, The Netherlands
        {\tt\small h.ayoobi@rug.nl}}%
        }
\usepackage{float}
\usepackage{graphicx}
\usepackage{comment}
\usepackage{amsmath,amssymb} % define this before the line numbering.
\usepackage{color}
\usepackage{algpseudocode}
\usepackage{caption}
\usepackage{subcaption}
\usepackage{mathtools}
\usepackage{caption}
\usepackage[
  separate-uncertainty = true,
  multi-part-units = repeat
]{siunitx}
\usepackage{subcaption}
\usepackage{float}

\DeclarePairedDelimiter\floor{\lfloor}{\rfloor}
\definecolor{red}{rgb}{1.00,0.00,0.00}
\definecolor{blue}{rgb}{0.00,0.00,1.00}
\definecolor{green}{rgb}{0.2,0.70,0.2}
\definecolor{yellow}{rgb}{0.5,0.5,0.0}

\usepackage{graphicx}
\usepackage{amsmath,amssymb,amsfonts}
\usepackage{hyperref}
\usepackage[ruled,vlined]{algorithm2e}
\usepackage{multirow}
\usepackage{tabularx, booktabs}
\usepackage{mathtools}
\usepackage{arydshln}
\usepackage{xcolor}
\usepackage{mwe}
\usepackage{wrapfig}
\usepackage[bottom]{footmisc} % put footnote under figures
\usepackage{hyperref}
\usepackage{makecell}
\usepackage{todonotes}
\begin{document}

\maketitle
\thispagestyle{empty}
\pagestyle{empty}

%%%%%%%%%%%%%%%%%%%%%%%%%%%%%%%%%%%%%%%%%%%%%%%%%%%%%%%%%%%%%%%%%%%%%%%%%%%%%%%%
\begin{abstract}

Local-HDP (for Local Hierarchical Dirichlet Process) is a hierarchical Bayesian method that has recently been used for open-ended 3D object category recognition. This method has been proven to be efficient in real-time robotic applications. However, the method is not robust to a high degree of occlusion. We address this limitation in two steps. First, we propose a novel semantic 3D object-parts segmentation method that has the flexibility of Local-HDP. This method is shown to be suitable for open-ended scenarios where the number of 3D objects or object parts is not fixed and can grow over time. We show that the proposed method has a higher percentage of mean intersection over union, using a smaller number of learning instances. Second,  we integrate this technique with a recently introduced argumentation-based online incremental learning method, thereby enabling the model to handle a high degree of occlusion. We show that the resulting model produces an explicit set of explanations for the 3D object category recognition task.  

\end{abstract}

%%%%%%%%%%%%%%%%%%%%%%%%%%%%%%%%%%%%%%%%%%%%%%%%%%%%%%%%%%%%%%%%%%%%%%%%%%%%%%%%
\section{INTRODUCTION}

Object  parts segmentation is one of the challenging problems in 3D shape analysis. Data-driven part-segmentation methods typically outperform traditional geometrical methods \cite{10.1145/2988458.2988473}. In recent years, deep learning approaches have been widely exploited among researchers in this field \cite{Yu_2019_CVPR}. Although these techniques show promising results in some applications, they are not well-suited for open-ended learning scenarios where the number of object categories and part-segments is  not predefined and can be extended over time.

The majority of existing models for 3D shape segmentation have the following five limitations that hamper their use in open-ended dynamic environments. First, most of these models are trained with a fixed set of labels, which greatly limits their flexibility and adaptivity. For instance, a model trained to segment a table into three semantic parts cannot be used to correctly segment a table with four parts. Second, using a fixed set of labels limits the number of object categories that the model can segment. Third, for state-of-the-art techniques, good learning accuracy requires a long training time. This prevents the model from quickly adapting to  changes in an open-ended dynamic environment.  Fourth, the object-parts segmentation and object category recognition methods in the literature typically use a large training set, while learning with a lower number of learning instances is required for quick adaptation of the model to changes. Fifth, 3D object category recognition techniques are typically not robust to a high degree of occlusion while encountering  occluded objects is common in real-world environments. These limitations motivated us to design an open-ended 3D object segmentation and recognition method for categorizing highly occluded objects in an open-ended manner with a lower number of learning instances. We thus make the following contributions: 

\begin{figure}[t!]

\centering
\includegraphics[width=1\linewidth]{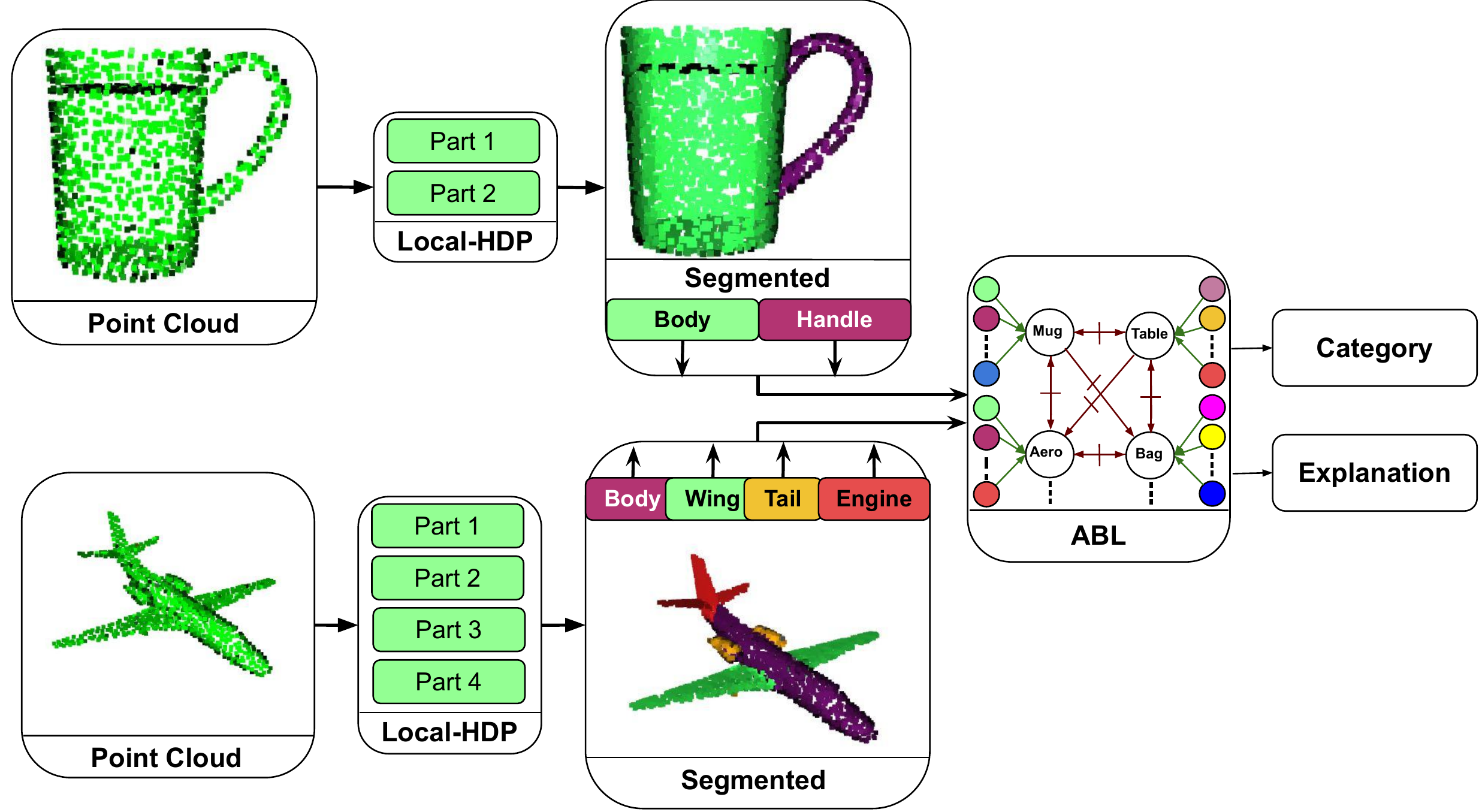}

\caption{The architecture of the proposed method. The proposed method is used for 3D object-parts segmentation. Subsequently, the argumentation-based learning method uses segmentation labels to recognize the 3D object categories and generate an explanation for the predicted categories. }
\label{fig:arc-category-recognition}
\end{figure}

\begin{enumerate}
    \item Introducing an open-ended 3D object-parts segmentation approach that extends the Local Hierarchical Dirichlet Process (Local-HDP) method \cite{Hamed_RAS_2021} and outperforms state-of-the-art segmentation techniques.
    % Firstly, it uses local-to-global and global-to-local object descriptors to represent each point in the point cloud of an object with respect to all the other points. 
    In contrast to Local-HDP, the proposed method does not require the construction of a dictionary. 
    % Here, each bin in the histogram of the descriptors is like a word in a dictionary. 
    This is important for open-ended applications since there is no need for pre-processing and the model can easily adapt to the changes in the environment.
    \item Proposing an open-ended 3D object category recognition technique based on Argumentation-Based Learning (ABL) \cite{Hamed_TASE_2021} that can handle a high degree of occlusion while generating explanations for the prediction of an object category. So far, ABL outperforms state-of-the-art methods in terms of learning accuracy and learning speed with low-dimensional datasets only, because of its high computational complexity \cite{hamedAABL2021}. The proposed approach enables the model to handle high-dimensional data as well. 
\end{enumerate}

Figure \ref{fig:arc-category-recognition} shows the architecture of the proposed model for segmenting two 3D objects, recognizing their corresponding categories, and generating explanations.

\section{Related Works}
3D object segmentation is a challenging task \cite{THEOLOGOU201549,Rodrigues2018}. 
% Methods of 3D object segmentation can be categorized based on their different characteristics, e.g, goal (surface type vs. part-type), geometry (region-based vs. boundary-based), type of learning (supervised vs. unsupervised), user involvement (automatic vs. interactive), number of 3D sensors used as information sources (single vs. multiple), type of features (geometric vs. structural), and finally the granularity of the produced result (hierarchical vs. non-hierarchical) \cite{THEOLOGOU201549}. In this research, we focus on non-hierarchical, part-type, supervised, interactive 3D object segmentation. 
% Our method is data-driven and can learn to segment new objects in an open-ended manner.
 State-of-the-art methods for 3D object-parts segmentation tasks include: PointNet~\cite{Qi_2017_CVPR}, PointNet++~\cite{NIPS2017_7095}, PointCNN~\cite{NIPS2018_7362}, O-CNN~\cite{10.1145/3072959.3073608}, SSCN~\cite{Graham_2018_CVPR}, PCNN~\cite{10.1145/3197517.3201301}, SPLATNet~\cite{Su_2018_CVPR} and PartNet~\cite{Yu_2019_CVPR}.  PartNet is a deep neural network shown to outperform all other approaches~\cite{Yu_2019_CVPR}. Unlike all these methods, our method is not based on deep neural networks and can be used in open-ended domains where the number of objects and the number of part-segments are not predetermined. This is not the case for the other approaches with a fixed number of class labels that require retraining the model when a new object category is added to the dataset. Furthermore, unlike deep neural architectures, our approach does not need a long separate training process before testing the model. This means that the model is trained incrementally and can be tested in real-time applications. This is particularly suitable for robotic applications. The large number of parameters in deep neural architectures requires a large number of training examples for optimization. However, experiments have shown that our model is capable of obtaining the same level of performance by only observing a small fraction of the training set used in other approaches based on deep learning. 
 
State-of-the-art approaches for the 3D object-segmentation task are typically tested with complete 3D objects that have no occlusion. However, in real-world (robotic) scenarios, it is common to have occluded objects in the scene and a method should also work well with occluded objects in the testing set. Our proposed method can perform well in the presence of a high degree of occlusion in the testing set, even if the model has been trained with complete 3D object models.  

Open-ended 3D object category recognition has been addressed in the literature \cite{kasaei2019local,Hamed_RAS_2021}. These methods typically use local models to achieve a class-incremental open-ended model. Although these models perform well in some real-world applications, they are typically not robust to a high degree of occlusion.

\section{Background}
In this section, the Local Hierarchical Dirichlet Process (Local-HDP) for 3D object category recognition \cite{Hamed_RAS_2021} is discussed in more detail. Furthermore, Argumentation-Based Learning (ABL) and Accelerated Argumentation-Based Learning (AABL) methods \cite{Hamed_TASE_2021,hamedAABL2021,hamed2019} are explained. 

\subsection{Local Hierarchical Dirichlet Process for 3D Object Category Recognition}
Local-HDP is a non-parametric hierarchical Bayesian model that has recently been introduced for 3D object category recognition in offline and open-ended scenarios. This method uses the non-parametric nature of the HDP \cite{Teh2006HDP2} together with the fast posterior approximation of the online variational inference technique \cite{wang2011online}. It constructs a local model for each object category.  For open-ended scenarios, the robotic system interacts with a human teacher or a simulated teacher in order to learn new object categories when needed and to extend the number of learned categories over time.  Studies have shown that Local-HDP learns with a lower number of  instances compared to state-of-the-art methods for 3D object category recognition. Local-HDP has higher learning accuracy and lower memory consumption than the compared methods \cite{kasaei2019local,Teh2006HDP2,blei2003latent}. Therefore, it is suitable for real-time open-ended robotic applications.  

\subsection{Argumentation-Based Learning}
Argumentation-Based Learning \cite{Hamed_TASE_2021,hamed2019} is an online incremental learning method that is based on argumentation theory and it is composed of two argumentation frameworks. Argumentation theory is a reasoning model that models the interaction between multiple arguments \cite{van2014handbook}. Dung defines an Abstract argumentation Framework (AF) as a pair consisting of a set of arguments and a binary relation showing the interaction between the arguments; this is the attack relation between two arguments. ABL assumes that the inner structure of the arguments A and B is $pre \rightarrow post$ where $pre$ is the reason for choosing a specific category $post$.  Therefore, A and B bidirectionally attack each other if and only if $A.pre = B.pre$ but $A.post \neq B.post$. ABL also uses a Bipolar Argumentation Framework (BAF) \cite{amgoud2008bipolarity} that has support relations in addition to the attack relations. An argument A supports B if and only if $A.post = B.pre$. ABL also defines an argument that is neither attacked nor supported as a supporting argument. 

% Argumentation-based learning uses a BAF unit for generating hypotheses from the learning instances and and an AF unit for modeling the interaction between the generated hypotheses in order to predict the class labels for the testing instances. 
ABL outperformed state-of-the-art online incremental learning techniques, neural architectures, deep reinforcement learning, and contextual bandit algorithms in the experimental results. This method is capable of learning with a lower number of learning instances while achieving a higher learning accuracy. However, ABL is limited to low-dimensional datasets due to its high computational complexity. This problem is addressed in Accelerated Argumentation Based Learning (AABL) technique \cite{hamedAABL2021}. AABL simplifies the model of the ABL %by using two mechanisms 
to reduce the computational time and space complexity of the method from exponential to polynomial. 
% Both the run-time and memory consumption of AABL are lower than those of ABL. Experimental results show that AABL also has better learning accuracy than ABL.

\section{Method}
The main goal of the proposed 3D object segmentation method is to independently learn topics for each semantic part-segment of the object using local-to-global and global-to-local shape descriptors. In this way, the input to the model is the description of a keypoint in an object and the output is the semantic label of that point. After segmenting a 3D object into a set of semantic parts, an argumentation-based learning approach \cite{Hamed_TASE_2021} is trained with the generated part labels in order to recognize the object categories.

\subsection{Local to Global 3D shape Descriptor}
\begin{figure}[b]

\centering
\includegraphics[width=0.95\linewidth]{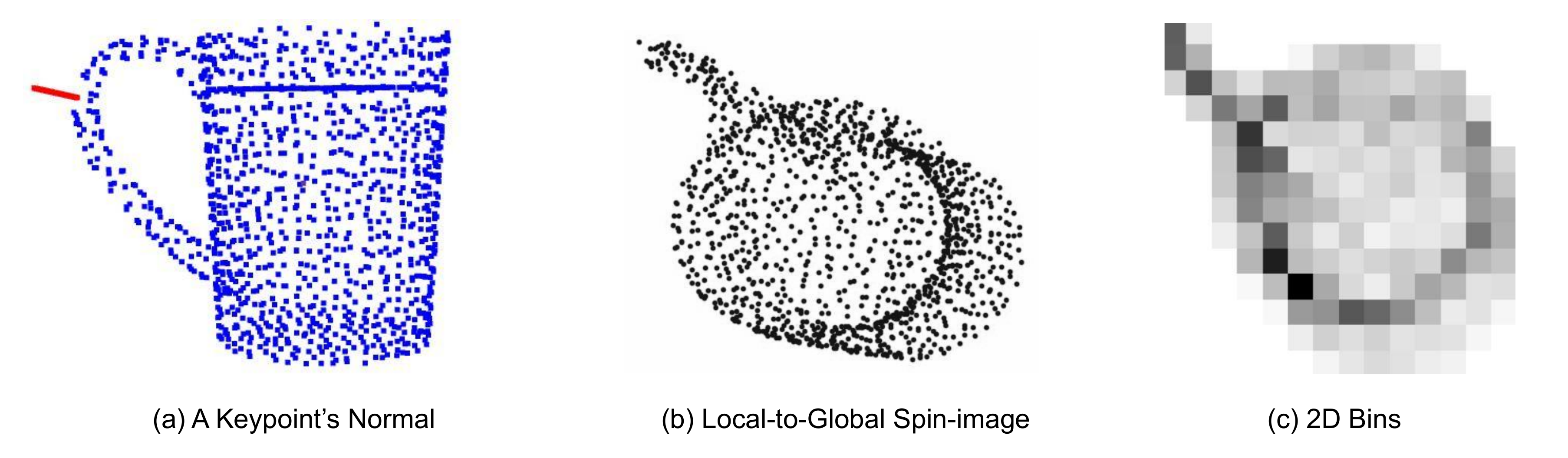}
\caption{a) The surface normal of a keypoint, shown in red. b) The corresponding local-to-global spin-image of the keypoint. c) The representation of the spin-image in 2D bins. }
\label{figspins}
\end{figure}

The input to the model is the BoWs representation of the mixture of local-to-global and global-to-local shape descriptors for each individual point in the point cloud.  This means that there is no need for dictionary construction anymore and each bin in the histogram of a descriptor is considered as a word in a dictionary.  The voxel grid method \footnote{\url{http://docs.pointclouds.org/trunk/classpcl_1_1_voxel_grid.html}} is used to  down-sample the point cloud. Spin images \cite{johnson1999using} have been used for the local-to-global representation of the points in the 3D objects. This means that the supporting radius of the spin-images is considered large enough to cover all the points in the object's point cloud. Figure \ref{figspins} shows the local-to-global spin-image of a point and its representation in 2D bins.  
% It is incorporated in combination with a global-to-local descriptor, which is explained in the next subsection, as inputs to the Local-HDP model. The obtained representation is then used to infer a set of topics in the topic layer.

\subsection{Global to Local 3D Object Descriptor}
  The global-to-local descriptor can pinpoint any point \textit{p} in the point cloud using a repeatable global reference frame. This descriptor helps the model find the location of each point on the point cloud. In order to find the global reference frame, we use the same methodology as the GOOD global 3D object descriptor \cite{kasaei2016good}. We use Principal Component Analysis (PCA) to find the three most dominant eigenvectors that show the principal directions with regard to the distribution of the points in the point cloud.  Like GOOD, for each projected point $\hat{p}=(\alpha, \beta)$ from the point cloud $P$, where $\alpha$ is along the $x$-axis and $\beta$ is along the $y$-axis, a row $r(\hat{p})$ and a column $c(\hat{p})$ are defined as follows:

\begin{equation}
    r(\hat{p}) = \floor{\frac{\alpha + \frac{l}{2}}{\frac{l+\epsilon}{n}}} \;\;\;\;
,\;
    c(\hat{p}) = \floor{\frac{\beta + \frac{l}{2}}{\frac{l+\epsilon}{n}}}
\end{equation}

\begin{figure}[t]
% \vspace{-10mm}
\centering
\includegraphics[width=0.95\linewidth]{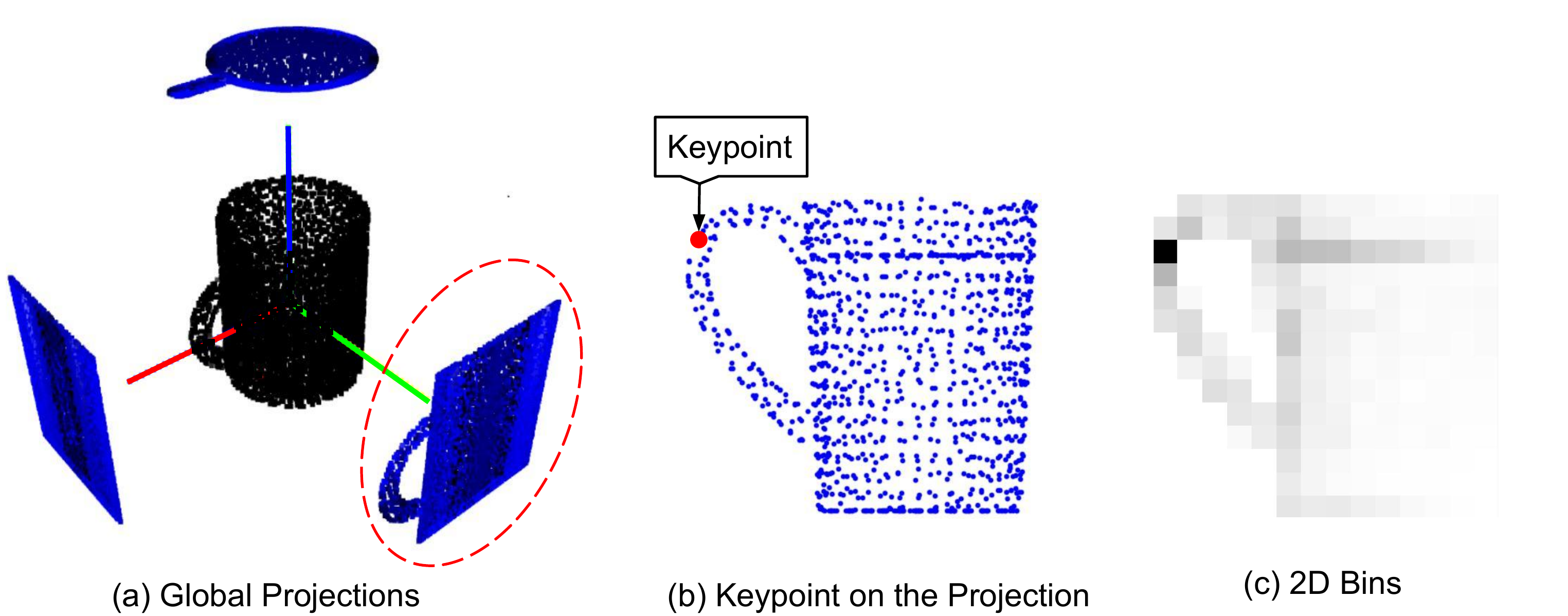}
\caption{a) The global projections of a mug object. b) A keypoint on a selected projection is shown with a red dot. c) The projected point-cloud is down-sampled to a set of bins and the keypoint is pinpointed on it. \vspace{-2mm}}
\label{fig:projections}
\end{figure}

\noindent where $l$ is the support length, $n$ is the number of bins and $\epsilon$ is a small value used for robustness. The left panel of Figure \ref{fig:projections} shows the three projected planes for a mug.  We use a different technique to pinpoint a point \textit{$p^*$} in the projection bins than the  global descriptor GOOD does. Namely, for each bin with row $i$ and column $j$, the value of each bin is calculated as follows:

\begin{equation}
    bin_{r,c}^{i,j} = \sum_{p \in (r(i),c(j))} (\frac{(l - d)}{l})^2
\label{eq_main}
\end{equation}

\noindent Here, $d=||p - p^*||$ is the Euclidean distance of point $p$  from the keypoint $p^*$ that we want to pinpoint. This process will be repeated for all the points in the point cloud and the three projection planes. At this step, we can find the location of each point in the point cloud of a 3D object that is not occluded using the aforementioned global-to-local descriptor. 

Notice that the local-to-global descriptor is more suitable for occluded objects since it only encodes the information in the local neighborhood of each point. Therefore,  the local-to-global descriptor is solely used in experiments with occluded objects.

\subsection{3D Object Segmentation using Local-HDP}
As stated previously, each point is represented as a set of descriptors with $V$ bins $\{s_0,\ldots,s_V\}$. For each object, we use Local-HDP to construct an independent local model for all the points located at each semantic part of the 3D object. Therefore, the number of local models corresponds to the number of learned object-parts.   

Local-HDP transforms the low-level spin-image features to conceptual high-level topics. Assuming that each point is a distribution over topics, each topic is itself a distribution over the bins of the descriptor ($s_i$). 
The topics in our method are only locally shared among the points of the same semantic segment of the object and not across all points. This means that the description of each word is only used to update its corresponding local model related to that part of the object. This uses an incremental inference approach where the numbers of object-parts and object categories are open-ended and not known beforehand. The proposed method also uses unsupervised HDP topic inference. However, a human teacher or simulation-based teacher can teach the model new objects and object-parts when a new object cannot be assigned to one of the previously learned categories.

After the model is constructed in a generative manner, the reverse procedure for inferring the latent variables from the data should be used. We have utilized the online variational inference technique method for the 3D object category recognition method. Instead of having a local model for each object category as in \cite{Hamed_RAS_2021}, here, each object-part corresponds to a local model.

\subsection{Local Online Variational Inference}
The posterior inference is intractable for HDP. Our approach adapted the online variational inference \cite{wang2011online} to approximate the inference method. This method is suitable for open-ended applications since it infers the topics in an online incremental manner.

A two-level Hierarchical Dirichlet Process (HDP) is made from a set of Dirichlet Processes (DP) sharing a basic distribution $G_0$. This distribution ($G_0$) is also drawn from a DP. Mathematically, this can be represented as follows: 

\begin{equation}
\begin{gathered}
    G_0 \sim DP({\gamma}H) \;\;\;\;\;,\;
    G_j \sim DP({\alpha_0}G_0)
\end{gathered}
\end{equation}

\noindent $G_j$ represents a DP for each document. All $G_j$s share the same atoms and the only difference between the $G_j$s is related to their corresponding weights. \textit{H} is a symmetric Dirichlet over the vocabulary simplex known as topics. 

To generate the data from HDP, first, the topic association for the $j$th document ($\theta_{jn}$) is drawn ($\theta_{jn} \sim G_j$). Second, the $n$th word of the $j$th document ($w_{jn}$) is generated using the topics ($w_{jn} \sim Mult(\theta_{jn})$).

\noindent An inference algorithm performs the reverse process by inferring the distributions over topics and the distribution over vocabulary words (${\theta}s$) using the dataset of documents. For the segmentation task, each document is the BoWs representation of the shape descriptors for each point from the point cloud.

Using Sethuraman’s stick-breaking construction for HDP \cite{Teh2006HDP2}, the variational distribution for online variational inference is as follows. 

\begin{equation}
    q(\beta', \pi', c, z, \phi) = q(\beta')q(\pi')q(c)q(z)q(\phi).
\end{equation}

\noindent Here, $\beta'=(\beta_{k}^{'})_{k=1}^{\infty}$ is the top-level stick proportion, $\pi'=(\phi_{jt}^{'})_{t=1}^{\infty}$ is the bottom-level stick proportion, and $c_j = (c_{jt}^{'})_{t=1}^{\infty}$ is the vector of indicators for each $G_j$. Moreover, $\phi = (\phi_k)_{k=1}^{\infty}$ is the inferred topic distribution, $z_{jn}$ is the topic index for each word $w_{jn}$.

The factorized form of  $q(c),\; q(z),\; q(\phi) ,\; q(\beta')\; \text{and}\; q(\pi')$ is the same as the online variation inference for HDP \cite{wang2011online}. Assuming that we have $|P|$ object-parts, the variational lower bound for document $j$ in the semantic object-part $P$ is calculated as follows:
\begin{multline}
    L_j^{(P)}= \mathop{{}\mathbb{E}_q}[log(p(w_j|c_j,z_j,\phi)p(c_j|\beta')p(z_j|\pi')p(\pi_j^{'}|\alpha_0))] \\ +   H(q(c_j))  + H(q(z_j)) +H(q(\phi')) \\ +  \frac{1}{|P|}[E_q[log p(\beta')p(\phi)] + H(q(\beta')) + H(q(\phi)]
\end{multline}
$H(.)$ is the entropy term for the variational distribution. For each 3D object-part, the lower bound is calculated in the following way:
\begin{equation}
    L^{(P)} = \sum_{j \in P} L_j = \mathop{{}\mathbb{E}_j}[|P|L_j]
\end{equation}

\noindent Using coordinate ascent equations \cite{pmlr-v15-wang11a} in the same way as online variation inference for HDP, the document-level parameters $(a_j, b_j, \psi_j, \zeta_j)$ and the category-level parameters $(\lambda^{(P)}, u^{(P)},v^{(P)})$ are inferred in the same way as Local-HDP \cite{Hamed_RAS_2021}. 

\begin{table*}

\scriptsize
\centering
\newcolumntype{?}{!{\vrule width 0.5pt}}
\setlength\arrayrulewidth{0.5pt}
\setlength\tabcolsep{0.5pt} % default value: 6pt
\renewcommand{\arraystretch}{1.2}% Tighter
\parbox{.7\linewidth}{
\begin{tabular}{  c  c  c  c  c  c  c  c  c  c  c  c  c  c  c  c  c}
\hline
 Method & \; Air\;\;\; & Bag\;\;\; & Cap\;\;\; & Car\;\; & Chair\;\;&  Eph. \;\;
 & Guitar\;\; & Knife\;\; & Lamp\;\;\; & Laptop \; & Motor \;\; & Mug\;\; 
 & Pistol\;\;
 & Rocket\;\; & Skate \;\; & Table \\
\hline

PN & 83.4 & 78.7 & 82.5 & 74.9 & 89.6 & 73.0
& 91.5 & 85.9 & 80.8 & 95.3 & 65.2 & 93.0 & 81.2 
& 57.9 & 72.8 & 80.6  \\ 

PN++ & 82.4 & 79.0 & 87.7 & 77.3 & 90.8 & 71.8 
& 91.0 & 85.9 & 83.7 & 95.3 & 71.6 & 94.1 & 81.3 
& 58.7 & 76.4 & 82.6   \\ 

O-CNN & 85.5 & 87.1 & 84.7 & 77.0 & 91.1 & 85.1 
& 91.9 & 87.4 & 83.3 & 95.4 & 56.9 & 96.2 & 81.6 
& 53.5 & 74.1 & 84.4  \\ 

SSCN  & 84.1 & 83.0 & 84.0 & 80.8 & 91.4 & 78.2 
& 91.6 & 89.1 & 85.0 & 95.8 & 73.7& 95.2 & 84.0 
&58.5 &76.0 &82.7  \\ 

PCNN  & 82.4 & 80.1 & 85.5 & 79.5 & 90.8 & 73.2 
& 91.3 & 86.0 & 85.0 & 95.7 & 73.2 & 94.8 & 83.3 
& 51.0 & 75.0 & 81.8  \\ 

SPLAT  & 83.2 & 84.3 & 89.1 & 80.3 & 90.7 & 75.5 
& 92.1 & 87.1 & 83.9 & 96.3 & 75.6 & 95.8 & 83.8 
& 64.0 & 75.5 & 81.8  \\ 

PtCNN  & 84.1 & 86.4 & 86.0 & 80.8 & 90.6 & 79.7 
& 92.3 & 88.4 & 85.3 & 96.1 & 77.2 & 95.3 & 84.2 
& 64.2 & 80.0 & 83.0  \\ 

PartNet & 87.8 & 86.7 & 89.7 & 80.5 & 91.9 & 75.7 
& 91.8 & 85.9 & 83.6 & 97.0 & 74.6 & 97.3 & 83.6 
& 64.6 & 78.4 & 85.8  \\ 
\hline
Our & \textbf{88.1} & \textbf{88.2} & \textbf{90.4} & \textbf{81.2} & \textbf{92.4} & \textbf{87.5} 
& \textbf{92.8} & \textbf{89.5} & \textbf{85.6} & \textbf{97.9} & \textbf{78.0} & \textbf{98.1} & \textbf{84.3} 
& \textbf{65.8} & \textbf{80.7} & \textbf{85.9}  \\ 
\hline
\end{tabular}
\vspace{1mm}
\caption{Results of offline evaluation of our approach compared with other state-of-the-art approaches. The words `Air' and 'Eph.' are used for the `Airplane' and 'Earphone' categories, respectively.}
\label{tbl2}}
\hfill
\parbox{.267\linewidth}{
\vspace{10mm}
  \centering
  \setlength\arrayrulewidth{0.1pt}
% \captionsetup{width=.9\linewidth}
% \setlength\arrayrulewidth{2pt}
\footnotesize
    \begin{tabular}{c c}
    \hline
    \textbf{Method} & \textbf{ Time (s)}\\
    \hline
    PointNet++ \cite{NIPS2017_7095} & 114912 \\ 
    
    PartNet \cite{Yu_2019_CVPR} & 161568    \\ 
   
    Our & \textbf{253} \\
    \hline
    \end{tabular}
    \vspace{10mm}
  \caption{Comparing the training time of two deep learning approaches with our approach.}
  \label{tbl:mot}}

\end{table*}

\section{Experimental Results}
Using a platform with Intel(R) Core(TM) i7, 3.30GHz processor, and 16GB RAM with SSD hard disk, we compared our proposed 3D semantic segmentation method using the adapted online variational inference technique with other methods. Specifically for open-ended evaluation, the comparison has been done with Local-LDA \cite{kasaei2019local}, LDA with shared topics \cite{blei2003latent}, BoWs \cite{kasaei2015adaptive}, RACE \cite{oliveira20163d}, and  HDP with shared topics and the online variational inference technique \cite{wang2011online}.
For offline evaluation of the method, we have compared our method with PointNet (PN) \cite{Qi_2017_CVPR}, PointNet++ (PN++) \cite{NIPS2017_7095}, PointCNN (PtCNN)~\cite{NIPS2018_7362}, O-CNN~\cite{10.1145/3072959.3073608}, SSCN~\cite{Graham_2018_CVPR}, PCNN~\cite{10.1145/3197517.3201301}, SPLATNet (SPLAT) \cite{Su_2018_CVPR} and PartNet~\cite{Yu_2019_CVPR}. 

% The experimental results for the 3D object-parts segmentation show that the combination of the two aforementioned local-to-global and global-to-local descriptors can lead to a better representation for each point in the point cloud.

We have conducted two sets of experiments to evaluate the performance of the proposed method. Firstly, we use the ShapeNet core segmentation dataset \cite{Mo_2019_CVPR} for offline evaluation of all the methods. This dataset consists of 16,881 shapes from 16 3D object categories, which contain 50 parts in total. Like all the other approaches, the mean Intersection over Union (mIoU)(\%) metric has been used for the evaluation of the part-based segmentation quality. 
 For the second set of experiments, the same dataset has been used. However, this time a simulated teacher introduced an object to an agent after random shuffling of all instances in the dataset.

\subsection{Offline Evaluation of the 3D Object-Parts Segmentation}
\begin{figure}[b]
% \vspace{-5mm}
\centering
\includegraphics[width=0.95\linewidth]{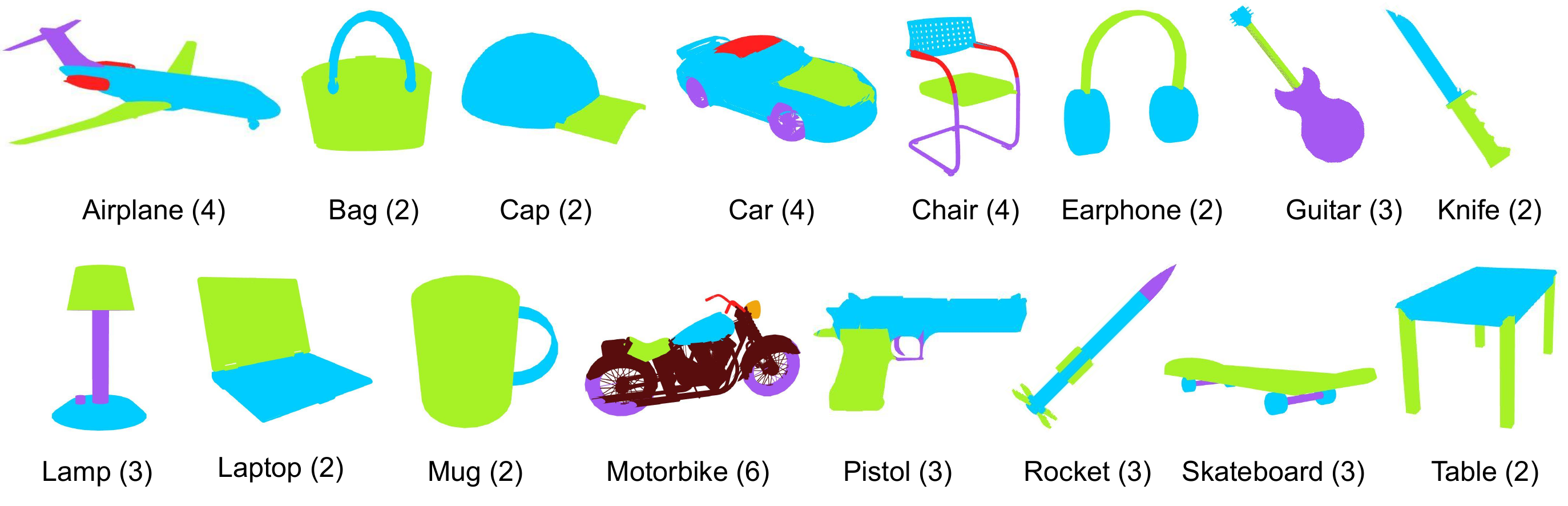}
\caption{Different object categories of the ShapeNet core dataset with 47 semantic object-parts. The numbers of object-parts are written between parentheses.}
\label{partnet}
\end{figure}

For offline evaluation of the proposed method, we used 10-fold cross-validation.  Figure \ref{partnet} shows some of the object categories of the Shapenet core dataset.

% \begin{figure}
% \centering
% \begin{minipage}{.45\textwidth}
%   \centering
%   \includegraphics[width=1\linewidth]{figures//Predictions.pdf}
%   \captionof{figure}{Examples of the resulting 3D object segmentation outputs using the proposed Local-HDP. The red points show the wrongly segmented points.}
%   \label{shapenet-predictions}
% \end{minipage}
% \end{figure}

\noindent Table \ref{tbl2} shows the comparison of offline evaluation of our approach versus other approaches. The proposed method for the objects parts segmentation task outperformed other approaches. 
%  Figure \ref{shapenet-predictions} shows the segmented 3D objects with false segmented points colored in red. This representation shows that most of the misclassified points are located near a region where two or more parts are connecting to each other. 
 Moreover, Table \ref{tbl:mot} shows the average training time for two well-known approaches in the literature compared to our proposed approach. 
 
 The reason for the higher mean segmentation intersection over union (mIoU \%) of the proposed approach compared to other state-of-the-art approaches lies in the combination of local-to-global and global-to-local object descriptors. Using the global-to-local shape descriptor can estimate the approximate location of each point in a 3D object. The local-to-global descriptor then leads to a more accurate estimation of each point's localization in the 3D point cloud. Moreover, using an independent local model for each 3D object-part leads to learning more details about the 3D structure of each object-part.

\subsection{Open-Ended Evaluation of the 3D Object-Parts Segmentation}
\iffalse  In order to evaluate our model in an open-ended learning scenario, we used the Washington RGBD dataset \cite{WashingtonRGBDDS}. This is a large dataset of 300 common household objects which are organized in 51 categories. The Washington RGBD dataset consists of 250,000 3D object views. In order to capture these object views, each object is placed on a round table and the Kinect-style camera captures the whole rotation of the object. For each object, this process is repeated with the camera mounted at three different heights to provide different object view angles with respect to the horizon \cite{WashingtonRGBDDS}. Figure \ref{figWashington} shows different categories of objects presented in the Washington RGBD Dataset. We should emphasize that all the experiments have been done using depth data and that color data were not used in our experiments. \fi

\iffalse
\begin{figure}
\centering
\includegraphics[width=.7\linewidth]{figures/3DSEG_Figures/WashingtonRGBD.png}
\caption{The RGB images for the objects of the different categories in the Washington RGBD dataset. Notice that, using only the depth data and not using color data, the recognition of some of the categories of these objects is a hard task even for a human.}
\label{figWashington}
\end{figure}
\fi

For open-ended evaluation of our approach for 3D object segmentation, we have two sets of experiments. The same approach as Local-LDA \cite{kasaei2019local} is used for the first experiments. We use a simulated teacher to teach a new object-part to the model or ask the model to segment an unforeseen object. In case of wrong segmentation, the simulated teacher sends correcting feedback to the model. 
For teaching a new object-part, the simulated teacher presents three randomly selected objects to the model. After this step, the model is tested with a set of random unforeseen objects with the previously learned object-parts. Subsequently, the average part-wise mIoU ($\%$) for the segmentation is computed. If the corresponding mIoU ($\%$) is higher than a certain threshold $T=0.75$, the simulated teacher will teach a new category to the model. If the learning accuracy does not exceed the threshold $T$ after a certain number of iterations (100 for our experiments), the teacher supposes that the agent cannot learn further and stops the experiment.  In order to calculate the part-wise mIoU, a sliding window of size $3n$, where $n$ is the number of learned object-parts, has been used. More details on the online evaluation protocol which has been used in our experiments can be found in \cite{kasaei2018coping}.

\begin{table}[t]
% \vspace{-5mm}
\scriptsize
\centering
\newcolumntype{?}{!{\vrule width 1pt}}
\setlength\arrayrulewidth{0.5pt}
\setlength\tabcolsep{1pt} % default value: 6pt
\renewcommand{\arraystretch}{1}% Tighter
\subfloat[Local-LDA]{
\begin{tabular}{  c  c c  c c }
\hline
 Exp\# & \#CI & \#LP & AIP & mIoU (\%)\\
\hline
1& \textbf{891} & 38 & 9.19 & 0.78  \\ 

2& 654 & 37 & 8.40 & 0.76  \\ 

3& 514 & 39 & \textbf{7.87} & 0.77  \\ 

4& 873 & 38 & 8.14 & \textbf{0.79}  \\ 

5& 579 & 37 & 9.04 & 0.76  \\ 

6& 605 & \textbf{40} & 7.91 & 0.77  \\ 

7& 849 & 36 & 9.20 & 0.75  \\ 

8& 550 & 34 & 8.32 & 0.78  \\ 

9& 560 & 39 & 7.99 & 0.76  \\ 

10& 564 & 37 & 8.92 & 0.77  \\ 
\hline
\makecell{\textbf{$Avg$} \\ \textbf{$\pm{std}$}} & \makecell{\textbf{$564$} \\ \textbf{$\pm{147}$}} & \makecell{\textbf{$37.50$}\\ \textbf{$\pm{1.71}$}} & \makecell{\textbf{$8.49$} \\\textbf{$\pm{0.53}$}} & \makecell{\textbf{$0.76$} \\ \textbf{$\pm{0.01}$}}  \\ 
\hline
\end{tabular}}
\quad
\subfloat[our]{
\begin{tabular}{  c  c c  c c }
\hline
 Exp\# & \#CI & \#LP & AIP & mIoU (\%)\\
\hline
1& 1179 & \textbf{47} & 3.31 & 0.89  \\ 

2& 1224 & \textbf{47} & 3.66 & 0.90  \\ 

3& 1241 & \textbf{47} & 3.54 & 0.89 \\ 

4& 1217 & \textbf{47} & \textbf{3.13} & 0.91  \\ 

5& 1123 & \textbf{47} & 3.23 & 0.87  \\ 

6& 1235 & \textbf{47} & 3.81 & 0.88  \\ 

7& 1252 & \textbf{47} & 3.29 & \textbf{0.92}  \\ 

8& 1115 & \textbf{47} & 3.72 & 0.89 \\ 

9 & \textbf{1291} & \textbf{47} & 3.73 & 0.90  \\ 

10 & 1232 & \textbf{47} & 3.01 & 0.91  \\ 
\hline
\makecell{\textbf{$Avg$} \\ \textbf{$\pm{std}$}} & \makecell{\textbf{$1210$} \\ \textbf{$\pm{55}$}} & \makecell{\textbf{$47$} \\ \textbf{$\pm{0}$}} & \makecell{\textbf{$3.44$} \\ \textbf{$\pm{0.28}$}} & \makecell{\textbf{$0.89$} \\ \textbf{$\pm{0.01}$}}  \\ 
\hline

\end{tabular}}

\caption{Summary of 10 experiments for the open-ended evaluation.\vspace{-4mm} }
\label{tbl3}
\end{table}

For the first set of experiments,  10 independent experiments have been carried out. This way, the random initialization of the models cannot abruptly change the results. Moreover, the order of presenting the objects to both models is kept the same using the simulated teacher. Several performance measures have been used to evaluate the open-ended learning capabilities of the methods, namely: (i) the number of Learned object-parts ({\tt{\#}}LP); (ii) the number of Correction Iterations ({\tt{\#}}CI) by the simulated user; (iii) the Average number of stored Instances per object-Part (AIP) ; (iv) Mean part-wise IoU (mIoU), the average part-wise IoU of each open-ended experiment. The left part of Table \ref{tbl3} shows the results of the experiments for Local-LDA using the aforementioned performance measures. The right part of Table \ref{tbl3} shows the performance of our proposed method.

According to Table \ref{tbl3}, the average number of learned object-parts for our approach is all 47 object-parts from the 16 categories, while Local-LDA learned $37.50 \pm 1.71$ object-parts on average. This means that our proposed technique can learn better than Local-LDA using the same dataset and learning protocol. Since online variational inference has been used for inference, the AIP is much lower for our approach with the average number of $3.44 \pm 0.28$ instances per object-part, which validates memory efficiency. On the other hand, Local-LDA needs to save on average $8.49 \pm 0.53$ instances per object-part. In terms of global segmentation mIoU, our method performs much better than Local-LDA.

\begin{figure}[t]
\centering
\begin{subfigure}{.33\linewidth}
  \centering
  \includegraphics[width=1\linewidth]{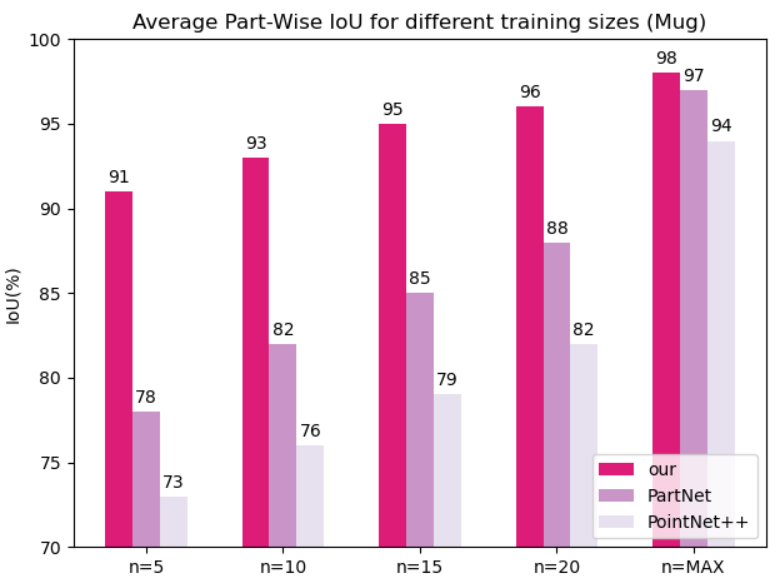}
  \caption{Mug}
  \label{mug_deep}
\end{subfigure}%
\begin{subfigure}{.33\linewidth}
  \centering
  \includegraphics[width=1\linewidth]{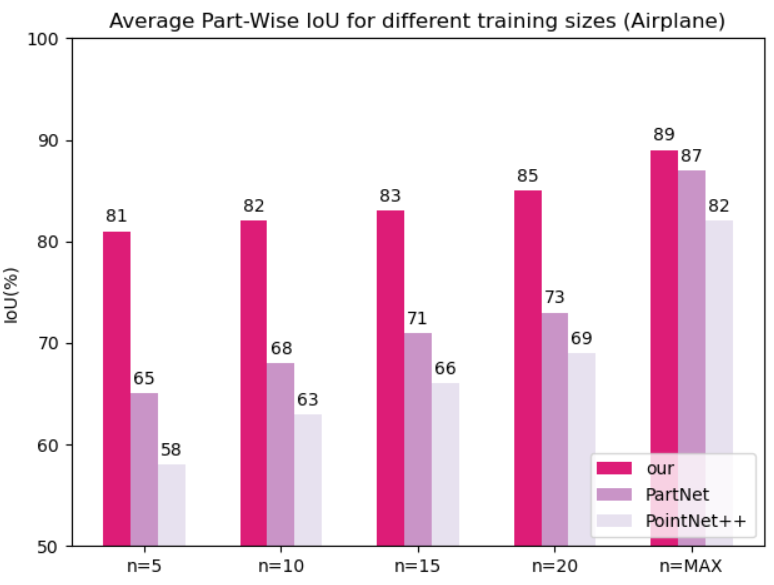}
  \caption{Airplane}
  \label{air_deep}
\end{subfigure}
\begin{subfigure}{.32\linewidth}
  \centering
  \includegraphics[width=1\linewidth]{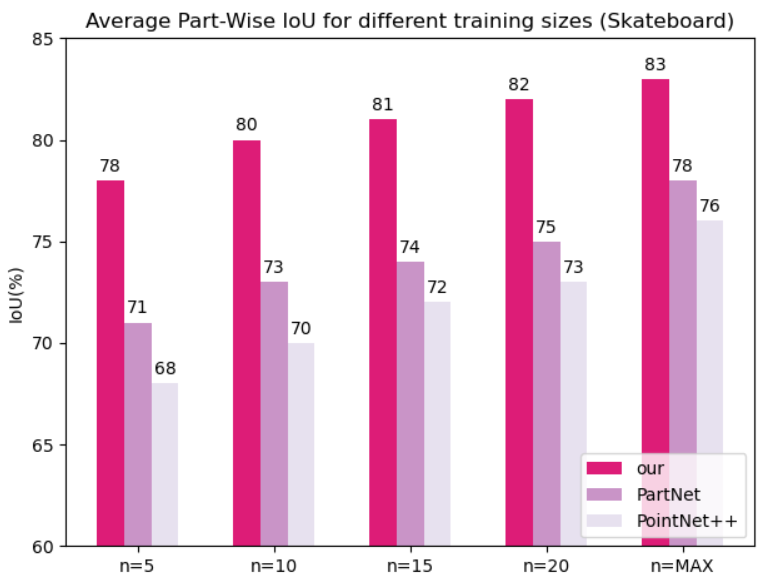}
  \caption{Skateboard}
  \label{skate_deep}
\end{subfigure}
\caption{The comparison of our proposed method for 3D object segmentation with PartNet and PointNet++ using four different training sizes ($n$). `Max' refers to the maximum size of a training set.\vspace{-4mm}}
\label{fig:barbar}

\end{figure}

For the second set of evaluations, our approach is compared to state-of-the-art deep architectures for 3D object segmentation (Figure \ref{fig:barbar}). For three randomly selected object categories, the segmentation mIoU(\%) is plotted using different numbers of training examples. A model with higher accuracy and a lower number of observations would then be more desirable for open-ended applications where the agent should quickly adapt to changes in the environment.

The results show that the proposed approach can generalize well with a small number of learning examples.

\subsection{Object Recognition with Argumentation-Based Learning}
At this step, the 3D objects are segmented into different parts and they are ready to be used as inputs to the Argumentation-Based online incremental Learning (ABL) method for predicting their categories. For a 3D object view $O$, the ABL model is trained over the segmented parts labels $(s_1, s_2,\ldots, s_n)$. 
% To evaluate the robustness of the proposed approach to occlusion, a simulated occluded dataset is constructed using the Shapenet core 3D objects. A randomly selected object has been occluded by randomly removing a piece of its point cloud (Figure \ref{fig:Random-cut}). 
To evaluate the robustness of different approaches to occlusion in the testing phase, the segmentation module is trained with the complete point clouds that have no occlusion and the occluded objects are only used in the testing set.  
% Providing the segmentation labels to ABL, the model predicts the category of an object and produces an explanation for choosing a specific category. 

\begin{figure}[b]
\centering
\includegraphics[width=.95\linewidth]{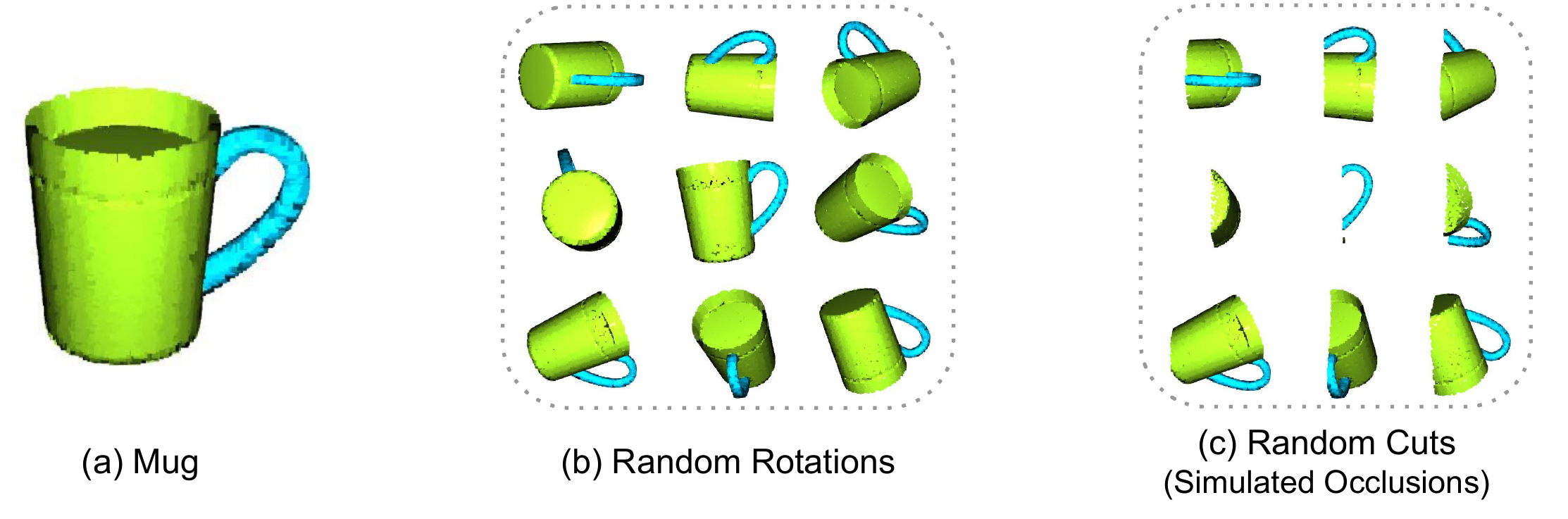}
\caption{The process of simulating the occluded objects. a) A mug. b) Random rotations of the selected mug. c) Random cuts from the random rotations.}
\label{fig:Random-cut}
\end{figure}

\subsubsection{Occlusion}
Using the Shapenet core dataset \cite{Mo_2019_CVPR}, a simulated occluded dataset is constructed. For this purpose, the objects are randomly rotated in 3D and a piece of its point-cloud is randomly removed using the following procedure. The minimum (min) and the maximum (max) values of the $x$-coordinates of all points in a point-cloud are used. To avoid removing too small or too large parts from the point-cloud of the rotated 3D object, a random cutting point on the $x$-axis is chosen in the range of $[min + \frac{max - min}{4}, max -  \frac{max - min}{4}]$.  Then, the points that have a lower $x$-coordinate value than the random cutting point are omitted. Figure \ref{fig:Random-cut} shows nine random rotations and random cuts. 
% This way a simulated occluded dataset is constructed for the testing set using the original Shapenet core dataset. 
% Although the occluded objects might form more shapes in real-world, the aforementioned process is chosen for simplicity.  

\begin{table}[t]
\scriptsize
\centering
\newcolumntype{?}{!{\vrule width 1pt}}
\setlength\arrayrulewidth{0.1pt}
\setlength\tabcolsep{2pt} % default value: 6pt
\renewcommand{\arraystretch}{1}% Tighter

\begin{tabular}{  c  c  c  c  c  c  c  c  c}
\hline 
 \multirow{2}{2em}{Ex\#} & \multicolumn{4}{c|}{Original Dataset}  & \multicolumn{4}{c}{Occluded Dataset} \\
\cline{2-9}
  &  Our & Local-HDP & PN++ & \multicolumn{1}{c|}{PrN}  & Our & Local-HDP & PN++ & PrN \\
\hline
1& 99 & 97 & 97 & 97& 83 & 45 & 21 & 25 \\ 

2&  97 &  96 & 96 & 96& 85 &  41 & 19 & 22\\ 

3&  98 & 97 & 96 & 97& 82 & 43 & 23 & 22 \\ 

4&  99 & 98 & 97 & 98& 87 & 38 & 21 & 24 \\ 

5&  98 & 98 & 97 & 97& 84 & 42 & 19 & 23  \\ 

6&  99 & 98 & 97 & 98&  83 & 47 & 22 & 24\\ 

7&  98 & 97 & 97 & 97&  88 & 40 & 21 & 25 \\ 

8&  99 & 98 & 96 & 98&  85 & 46 & 24 & 23\\ 

9&  99 & 98 & 97 & 97&  84 & 44 & 17 & 25\\ 

10&  98 & 97 & 97 & 97 & 84 & 46 & 19 & 24\\ 
\hline
\makecell{\textbf{$Avg$} \\ \textbf{$\pm{std}$}} & \makecell{\textbf{$98$} \\ \textbf{$\pm{0.69}$}} &  \makecell{\textbf{$97$} \\ \textbf{$\pm{0.69}$}} & \makecell{\textbf{$96$} \\ \textbf{$\pm{0.48}$}} & \makecell{\textbf{$97$} \\ \textbf{$\pm{0.63}$}} & \makecell{\textbf{$84$} \\ \textbf{$\pm{1.84}$}} & 
\makecell{$43$ \\ $\pm{2.93}$} & 
\makecell{$20$ \\ $\pm{2.11}$} & 
\makecell{$23$ \\ $\pm{1.15}$}\\ 
\hline
\end{tabular}
\caption{The comparison between the recognition accuracy (\%) of different approaches using the original dataset and the occluded dataset. Notice that the occluded dataset is only used for testing, not training.\vspace{-2mm}}
\label{tbl-occluded}
\end{table}

\begin{figure}[t]
\centering
\includegraphics[width=0.95\linewidth]{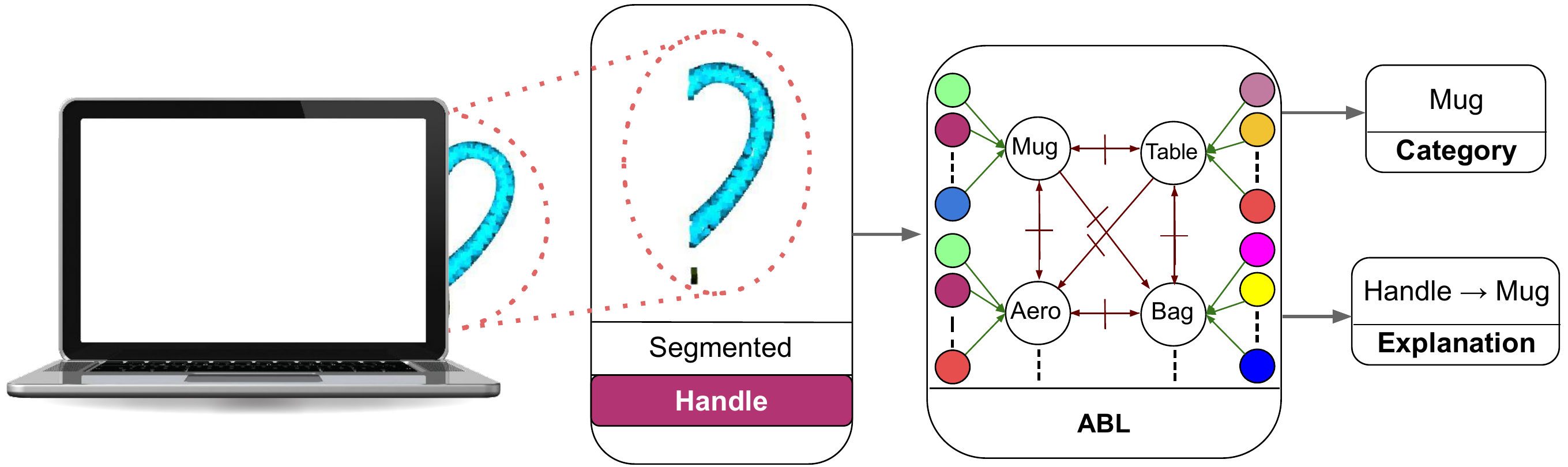}
\caption{An example of the generated explanations. An occluded mug's handle is first segmented. Subsequently, its category is predicted and explained. Here, $Handle \rightarrow Mug$ means that the reason for predicting the mug category is that the handle  looks like a typical mug handle.}
\label{fig:occluded explanations}
\end{figure}

To compare the performance of different approaches on the original Shapenet core and the occluded dataset, two sets of experiments are conducted. The proposed approach is compared with the Local-HDP \cite{Hamed_RAS_2021}, PointNet++ (PN) \cite{NIPS2017_7095} and PartNet (PrN++) \cite{Yu_2019_CVPR}. The first set of experiments uses the original ShapeNet core dataset for both the training and the testing phase. However, the second set of experiments uses the occluded dataset in the testing set to evaluate the robustness of the model to occlusion. This means that in the second set of experiments, the model is trained using $90\%$ of the original Shapenet core dataset and the rest of the $10\%$ objects are occluded, using the aforementioned procedure, to construct the testing set. Notice that the 10-fold cross-validation technique is used for both sets of experiments and the resulting accuracies are reported for each fold.

Table \ref{tbl-occluded} summarizes the recognition accuracy (\%) of the trained model for different testing sets, namely, the original dataset and the occluded dataset. By comparing the obtained results, it is clear that the proposed method outperforms all the other methods. The results of recognition accuracies on the occluded dataset show that the proposed method is more robust to occlusion,  with an average $84\%$ accuracy. The other approaches, namely, Local-HDP, PointNet++, and PartNet, achieved  $43\%$, $20\%$, and $23\%$ learning accuracies, respectively. The proposed approach achieved on average $41\%$ higher accuracy than the second-best performing approach, namely, Local-HDP.      
Figure \ref{fig:occluded explanations} shows an example of an explanation generated by ABL for the object category classification task.

% \subsubsection{Unforeseen Categories}
% In order to evaluate the precision of the model for unforeseen categories with the same object-parts, the model is trained on the Shapenet core dataset \cite{chang2015shapenet} and the testing phase is done with the Washington RGBD dataset \cite{WashingtonRGBDDS}. The Washington RGBD dataset has 51 categories. The objects categories of with visual similarities to the training set are selected for the testing phase. Figure \ref{fig:new_category} show a teapot which has the same handle part as the learned mug. This way the segmentation module can recognize the parts of the new categories. Since, the new objects are not previously seen by the model, the prediction should be wrong. Therefore, a correcting feedback can explain to the ABL model why a new object should be classified in a different category. 

\section{Conclusion}
In this paper, a non-parametric hierarchical Bayesian method is proposed for 3D object-part segmentation. It is integrated with the Argumentation-Based online incremental Learning (ABL) technique for the 3D object category recognition task.
State-of-the-art techniques for 3D object-parts segmentation typically use deep neural networks that take a long time for training. Therefore, they are not suitable for open-ended or class-incremental scenarios where the number of class labels may increase over time. These methods typically require a large dataset to achieve a high learning accuracy. In contrast, the proposed model can learn with a lower number of learning instances while achieving higher part segmentation mean Intersection over Union (mIoU).  

The experimental results show that the proposed integrated model outperforms state-of-the-art methods in terms of accuracy, recognition accuracy, segmentation mIoU, and training time. Evaluating the robustness of different 3D object category recognition techniques to occlusion shows that the proposed technique outperforms other techniques by a large margin. Moreover, ABL provides an explanation for the 3D object category recognition task. These explanations seem suitable for human-robot interaction since a user can understand the underlying reasoning process and (s)he can debug the model by injecting a new set of explanations (arguments) into the model.

% \section*{ACKNOWLEDGMENT}
% Omitted for blind reviewing
% \noindent This work is conducted at the center of Data Science and Systems Complexity (DSSC) and sponsored by a  Marie Sk{\l}odowska-Curie COFUND grant, agreement no. 754315.

\bibliography{references}

\begin{thebibliography}{10}
\providecommand{\url}[1]{#1}
\csname url@samestyle\endcsname
\providecommand{\newblock}{\relax}
\providecommand{\bibinfo}[2]{#2}
\providecommand{\BIBentrySTDinterwordspacing}{\spaceskip=0pt\relax}
\providecommand{\BIBentryALTinterwordstretchfactor}{4}
\providecommand{\BIBentryALTinterwordspacing}{\spaceskip=\fontdimen2\font plus
\BIBentryALTinterwordstretchfactor\fontdimen3\font minus
  \fontdimen4\font\relax}
\providecommand{\BIBforeignlanguage}[2]{{%
\expandafter\ifx\csname l@#1\endcsname\relax
\typeout{** WARNING: IEEEtran.bst: No hyphenation pattern has been}%
\typeout{** loaded for the language `#1'. Using the pattern for}%
\typeout{** the default language instead.}%
\else
\language=\csname l@#1\endcsname
\fi
#2}}
\providecommand{\BIBdecl}{\relax}
\BIBdecl

\bibitem{10.1145/2988458.2988473}
K.~Xu, V.~G. Kim, Q.~Huang, N.~Mitra, and E.~Kalogerakis, ``Data-driven shape
  analysis and processing,'' in \emph{SIGGRAPH ASIA 2016 Courses}, ser. SA
  ’16.\hskip 1em plus 0.5em minus 0.4em\relax New York, NY, USA: Association
  for Computing Machinery, 2016.

\bibitem{Yu_2019_CVPR}
F.~Yu, K.~Liu, Y.~Zhang, C.~Zhu, and K.~Xu, ``Partnet: A recursive part
  decomposition network for fine-grained and hierarchical shape segmentation,''
  in \emph{The IEEE Conference on Computer Vision and Pattern Recognition
  (CVPR)}, June 2019.

\bibitem{Hamed_RAS_2021}
H.~Ayoobi, H.~Kasaei, M.~Cao, R.~Verbrugge, and B.~Verheij, ``Local-hdp:
  Interactive open-ended 3d object category recognition in real-time robotic
  scenarios,'' \emph{Robotics and Autonomous Systems}, vol. 147, p. 103911,
  2022.

\bibitem{Hamed_TASE_2021}
H.~Ayoobi, M.~Cao, R.~Verbrugge, and B.~Verheij, ``Argumentation-based online
  incremental learning,'' \emph{IEEE Transactions on Automation Science and
  Engineering}, pp. 1--15, 2021.

\bibitem{hamedAABL2021}
H.~Ayoobi, M.~Cao, R.~L.~C. Verbrugge, and B.~Verheij, ``Argue to learn:
  Accelerated argumentation-based learning,'' in \emph{20th IEEE International
  Conference on Machine Learning and Applications (ICMLA)}.\hskip 1em plus
  0.5em minus 0.4em\relax IEEE, 2021.

\bibitem{THEOLOGOU201549}
P.~Theologou, I.~Pratikakis, and T.~Theoharis, ``A comprehensive overview of
  methodologies and performance evaluation frameworks in {3D} mesh
  segmentation,'' \emph{Computer Vision and Image Understanding}, vol. 135, pp.
  49 -- 82, 2015.

\bibitem{Rodrigues2018}
R.~S.~V. Rodrigues, J.~F.~M. Morgado, and A.~J.~P. Gomes, ``Part-based mesh
  segmentation: A survey,'' \emph{Computer Graphics Forum}, vol.~37, no.~6, pp.
  235--274, 2018.

\bibitem{Qi_2017_CVPR}
C.~R. Qi, H.~Su, K.~Mo, and L.~J. Guibas, ``{PointNet}: Deep learning on point
  sets for {3D} classification and segmentation,'' in \emph{Proceedings of the
  IEEE Conference on Computer Vision and Pattern Recognition (CVPR)}, July
  2017.

\bibitem{NIPS2017_7095}
C.~R. Qi, L.~Yi, H.~Su, and L.~J. Guibas, ``{PointNet++}: Deep hierarchical
  feature learning on point sets in a metric space,'' in \emph{Advances in
  Neural Information Processing Systems 30}, I.~Guyon, U.~V. Luxburg,
  S.~Bengio, H.~Wallach, R.~Fergus, S.~Vishwanathan, and R.~Garnett, Eds.\hskip
  1em plus 0.5em minus 0.4em\relax Curran Associates, Inc., 2017, pp.
  5099--5108.

\bibitem{NIPS2018_7362}
Y.~Li, R.~Bu, M.~Sun, W.~Wu, X.~Di, and B.~Chen, ``Pointcnn: Convolution on
  x-transformed points,'' in \emph{Advances in Neural Information Processing
  Systems 31}, S.~Bengio, H.~Wallach, H.~Larochelle, K.~Grauman,
  N.~Cesa-Bianchi, and R.~Garnett, Eds.\hskip 1em plus 0.5em minus 0.4em\relax
  Curran Associates, Inc., 2018, pp. 820--830.

\bibitem{10.1145/3072959.3073608}
P.-S. Wang, Y.~Liu, Y.-X. Guo, C.-Y. Sun, and X.~Tong, ``O-cnn: Octree-based
  convolutional neural networks for {3D} shape analysis,'' \emph{ACM Trans.
  Graph.}, vol.~36, no.~4, Jul. 2017.

\bibitem{Graham_2018_CVPR}
B.~Graham, M.~Engelcke, and L.~van~der Maaten, ``{3D} semantic segmentation
  with submanifold sparse convolutional networks,'' in \emph{The IEEE
  Conference on Computer Vision and Pattern Recognition (CVPR)}, June 2018.

\bibitem{10.1145/3197517.3201301}
M.~Atzmon, H.~Maron, and Y.~Lipman, ``Point convolutional neural networks by
  extension operators,'' \emph{ACM Trans. Graph.}, vol.~37, no.~4, Jul. 2018.

\bibitem{Su_2018_CVPR}
H.~Su, V.~Jampani, D.~Sun, S.~Maji, E.~Kalogerakis, M.-H. Yang, and J.~Kautz,
  ``Splatnet: Sparse lattice networks for point cloud processing,'' in
  \emph{The IEEE Conference on Computer Vision and Pattern Recognition (CVPR)},
  June 2018.

\bibitem{kasaei2019local}
S.~H. Kasaei, L.~S. Lopes, and A.~M. Tomé, ``Local-lda: Open-ended learning of
  latent topics for 3d object recognition,'' \emph{IEEE Transactions on Pattern
  Analysis and Machine Intelligence}, vol.~42, no.~10, pp. 2567--2580, 2020.

\bibitem{hamed2019}
H.~{Ayoobi}, M.~{Cao}, R.~{Verbrugge}, and B.~{Verheij}, ``Handling unforeseen
  failures using argumentation-based learning,'' in \emph{2019 IEEE 15th
  International Conference on Automation Science and Engineering (CASE)}, Aug
  2019, pp. 1699--1704.

\bibitem{Teh2006HDP2}
Y.~Teh, M.~Jordan, M.~Beal, and D.~Blei, ``Sharing clusters among related
  groups: Hierarchical {Dirichlet} processes,'' \emph{Advances in neural
  information processing systems}, vol.~17, 2004.

\bibitem{wang2011online}
C.~Wang, J.~Paisley, and D.~Blei, ``Online variational inference for the
  hierarchical {Dirichlet} process,'' in \emph{Proceedings of the Fourteenth
  International Conference on Artificial Intelligence and Statistics}, ser.
  Proceedings of Machine Learning Research, G.~Gordon, D.~Dunson, and
  M.~Dudík, Eds., vol.~15.\hskip 1em plus 0.5em minus 0.4em\relax Fort
  Lauderdale, FL, USA: PMLR, 11--13 Apr 2011, pp. 752--760.

\bibitem{blei2003latent}
D.~M. Blei, A.~Y. Ng, and M.~I. Jordan, ``{Latent {Dirichlet} Allocation},''
  \emph{Journal of Machine Learning Research}, vol.~3, no. Jan, pp. 993--1022,
  2003.

\bibitem{van2014handbook}
F.~H. Van~Eemeren, B.~Garssen, E.~C. Krabbe, A.~F.~S. Henkemans, B.~Verheij,
  and J.~H. Wagemans, \emph{Handbook of argumentation theory}.\hskip 1em plus
  0.5em minus 0.4em\relax Dordrecht: Springer, 2014.

\bibitem{amgoud2008bipolarity}
L.~Amgoud, C.~Cayrol, M.-C. Lagasquie-Schiex, and P.~Livet, ``On bipolarity in
  argumentation frameworks,'' \emph{International Journal of Intelligent
  Systems}, vol.~23, no.~10, pp. 1062--1093, 2008.

\bibitem{johnson1999using}
A.~E. Johnson and M.~Hebert, ``{Using spin images for efficient object
  recognition in cluttered {3D} scenes},'' \emph{IEEE Transactions on Pattern
  Analysis and Machine Intelligence}, vol.~21, no.~5, pp. 433--449, 1999.

\bibitem{kasaei2016good}
S.~H. Kasaei, A.~M. Tom{\'e}, L.~S. Lopes, and M.~Oliveira, ``{GOOD: A global
  orthographic object descriptor for {3D} object recognition and
  manipulation},'' \emph{Pattern Recognition Letters}, vol.~83, pp. 312--320,
  2016.

\bibitem{pmlr-v15-wang11a}
C.~Wang, J.~Paisley, and D.~Blei, ``Online variational inference for the
  hierarchical {Dirichlet} process,'' in \emph{Proceedings of the Fourteenth
  International Conference on Artificial Intelligence and Statistics}, ser.
  Proceedings of Machine Learning Research, G.~Gordon, D.~Dunson, and
  M.~Dudík, Eds., vol.~15.\hskip 1em plus 0.5em minus 0.4em\relax Fort
  Lauderdale, FL, USA: PMLR, 11--13 Apr 2011, pp. 752--760.

\bibitem{kasaei2015adaptive}
S.~H. Kasaei, M.~Oliveira, G.~H. Lim, L.~S. Lopes, and A.~M. Tom{\'e}, ``{An
  adaptive object perception system based on environment exploration and
  Bayesian learning},'' in \emph{2015 IEEE International Conference on
  Autonomous Robot Systems and Competitions}.\hskip 1em plus 0.5em minus
  0.4em\relax IEEE, 2015, pp. 221--226.

\bibitem{oliveira20163d}
M.~Oliveira, L.~S. Lopes, G.~H. Lim, S.~H. Kasaei, A.~M. Tom{\'e}, and
  A.~Chauhan, ``{3D object perception and perceptual learning in the RACE
  project},'' \emph{Robotics and Autonomous Systems}, vol.~75, pp. 614--626,
  2016.

\bibitem{Mo_2019_CVPR}
K.~Mo, S.~Zhu, A.~X. Chang, L.~Yi, S.~Tripathi, L.~J. Guibas, and H.~Su,
  ``Partnet: A large-scale benchmark for fine-grained and hierarchical
  part-level {3D} object understanding,'' in \emph{The IEEE Conference on
  Computer Vision and Pattern Recognition (CVPR)}, June 2019.

\bibitem{kasaei2018coping}
S.~H. Kasaei, L.~S. Lopes, and A.~M. Tom{\'e}, ``Coping with context change in
  open-ended object recognition without explicit context information,'' in
  \emph{2018 IEEE/RSJ International Conference on Intelligent Robots and
  Systems (IROS)}.\hskip 1em plus 0.5em minus 0.4em\relax IEEE, 2018, pp. 1--7.

\end{thebibliography}
\bibliographystyle{IEEEtran}

\end{document}